\documentclass[conference]{IEEEtran}
\IEEEoverridecommandlockouts

\usepackage{bm}
\usepackage{threeparttable}
\usepackage{booktabs}
\usepackage{multicol}
\usepackage{multirow}
\usepackage{amsmath}
\usepackage{amsthm}
\usepackage{amssymb}
\usepackage{fancyhdr}
\usepackage{amsfonts}
\usepackage{ifpdf}

\usepackage{cite}

\ifCLASSINFOpdf
   \usepackage[pdftex]{graphicx}
\else
\fi
\fancypagestyle{plain}{
  \fancyhf{}
}
\usepackage{eso-pic}

\IEEEpubid{\makebox[\columnwidth]{2375-9259/20/\$31.00 \textcopyright 2020 IEEE. \hfill} \hspace{\columnsep}\makebox[\columnwidth]{ }}

\begin{document}
\AddToShipoutPictureBG*{%
  \AtPageUpperLeft{%
    \setlength\unitlength{1in}%
    \hspace*{\dimexpr0.5\paperwidth\relax}
    \makebox(0,-0.75)[c]{2020 International Conference on Data Mining Workshops (ICDMW)}%
}}
%
\title{Individualized Context-Aware Tensor Factorization for Online Games Predictions}
        


%
\author{\IEEEauthorblockN{
    Julie Jiang\IEEEauthorrefmark{1}\IEEEauthorrefmark{2},
    Kristina Lerman\IEEEauthorrefmark{1}\IEEEauthorrefmark{2},
    Emilio Ferrara\IEEEauthorrefmark{1}\IEEEauthorrefmark{2}\IEEEauthorrefmark{3}}
\IEEEauthorblockA{\IEEEauthorrefmark{1}
    Information Sciences Institute, University of Southern California, Marina del Rey, CA, 90292}
\IEEEauthorblockA{\IEEEauthorrefmark{2}
    Department of Computer Science, University of Southern California, Los Angeles, CA, 90089}
\IEEEauthorblockA{\IEEEauthorrefmark{3}
    Annenberg School of Communication, University of Southern California, Los Angeles, CA, 90089}}


%

\maketitle
\begin{abstract}
Individual behavior and decisions are substantially influenced by their contexts, such as location, environment, and time. Changes along these dimensions can be readily observed in Multiplayer Online Battle Arena games (MOBA), where players face different in-game settings for each match and are subject to frequent game patches. Existing methods utilizing contextual information generalize the effect of a context over the entire population, but contextual information tailored to each individual can be more effective. To achieve this, we present the Neural Individualized Context-aware Embeddings (NICE) model for predicting user performance and game outcomes. Our proposed method identifies individual behavioral differences in different contexts by learning latent representations of users and contexts through non-negative tensor factorization. Using a dataset from the MOBA game \textit{League of Legends}, we demonstrate that our model substantially improves the prediction of winning outcome, individual user performance, and user engagement.
\end{abstract}

\section{Introduction}
Understanding human behavior through data is a hot area of research in machine learning. Most data-driven methods generalize known behavioral patterns to unknown ones, typically framing the problem as an unsupervised or supervised prediction task. However, the heterogeneity of human behavior and contextual factors complicates the data formulation. In this paper, we consider the multiplayer online battle arena (MOBA) game League of Legends (LoL),  where the multitudes of human interaction and decision-making make it a mirror image of society in real life. The growing interest in online games and the wealth of available gameplay data have raised the possibility of modeling player performance through a data-driven approach. An LoL game is characterized as one standalone match, wherein two teams of players compete by trying to destroy the opposing team's base first. At the start of each match, each player assumes a role by controlling a champion as part of their attack or defense strategy, which in turn is largely dependent on the capabilities of their chosen champion.

Despite constraints in the gameplay environment, predicting the outcomes of online games is challenging due to variability in player skills and the changing game contexts. Individual players differ in their skill level, their mastery of a specific champion style, their game style preference, and their adaptability to different roles and situations. Even the hardware they play the game or the time lag due to poor internet connection and the geographical region could be partly responsible for their performances. In addition, game context changes globally as a function of the game version, both at the team level, based on the queue or tournament types they play in, and at the champion level, based on the upgrades or downgrades of a champion's skills and abilities.

To address these challenges, we develop NICE (Neural Individualized Context-Aware Embeddings) to enable user-context modeling that is applicable to many user behavior datasets. The underlying technique we employ is tensor factorization, which has emerged as a powerful tool for unsupervised, multi-modal analysis in high dimensional data \cite{kolda2009tensor}. Being a generalization of matrix factorization, tensor factorization enables n-way analyses of multi-linear data that better captures their structural representations. The main challenge limiting application-specific tensor factorization surrounds constructing valid tensors from high-order data. This is especially true for the LoL dataset given the large number of users, the varying numbers of matches played by each user, and the varying time points at which each user chooses to play. In this paper, we describe a generalized approach to constructing and using tensor factorization methods. In particular, we demonstrate that NICE works well in the case of multiple types of contexts, such as contexts that apply globally to all users and contexts that apply separately to each individual. By considering the population as a whole in our tensor, the learned latent factors serve to elucidate meaningful behavioral patterns specific to each individual. Finally, we show that the latent factors are suitable for various downstream applications via a deep neural network decoder.

A unique advantage of NICE is that it easily combines both high predictive power and high interpretability in a contextually aware setting. We evaluate our approach by predicting match outcomes and player performance in tens of thousands of LoL players and nearly half a million matches. We demonstrate how our model can be effectively utilized in a wide range of applications, from anticipating outcomes to estimating performance to predicting user engagement, and it systematically outperforms the baseline models. We also provide an in-depth explanation of the latent factors we reveal, which offers insights into the heterogeneity of users and the effects of various contextual settings. 

\begin{figure*}[t]
    \centering
    \includegraphics[width=\linewidth]{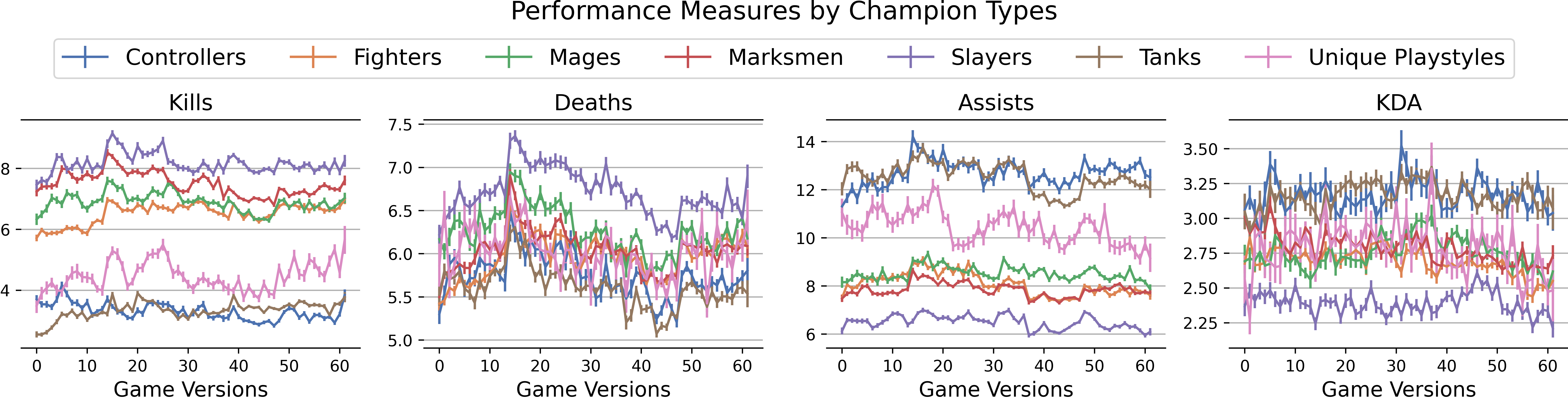}
    \caption{A line plot with error bars of the average number of \textit{kills}, \textit{deaths}, \textit{assists}, and \textit{KDA},attained by champions in each of the seven champion types across game versions. The error bars show the standard error of the mean.}
    \label{fig:champ_plots}
\end{figure*}

\section{Related Work}

Contextual awareness is important in many machine learning applications. The ability to understand the context and behave accordingly is as useful and crucial as model interpretability and generalizability, two topics that attract much attention. Although there exists research on promoting contextual-awareness in machine learning models, a primary limitation of current methods is their inability to generalize to multiple and categorical context variables, of which there are many in the real world. One direct approach is to model one type of context, e.g., time, as an additional dimension in a collaborative filtering based method \cite{biancalana2011context}. In another example, \cite{nascimento2018context} developed a localized context-aware method by training separate models for each context. However, these methods do not naturally generalize to cases of multiple contextual factors.

We approach contextualization using tensor factorization, for which there exists a broad body of research focusing on its analytical potentials and applications. The interpretability of factorization methods has enabled its adoption to understand human behavior \cite{sapienza2018non,hosseinmardi2018tensor,hosseinmardi2019discovering} and analyze complex temporal activities or signals \cite{dunlavy2011temporal,sapienza2015detecting}. We also draw inspiration from tensor factorization applications such as e-mail topic tracking \cite{bader2008discussion}, and patient clustering \cite{ruffini2017clustering}. Pertaining to game-related studies, \cite{sapienza2018non} mined user behavioral patterns in LoL using tensor factorization by measuring observable in-game performances.

The majority of MOBA game research has been focused on predicting factors influencing the outcome. Some tackles the problem by finding the most optimal team composition to maximize the chance of winning. For example, \cite{SapienzaTeamComposition} explores best teammate combinations and \cite{hanke2017recommender, semenov2016performance} examined the most advantageous hero (ie. champions) lineups in \textit{Dota 2}. \cite{Lan2018APB} used neural networks to extract player behavioral patterns to predict match outcomes. These methods do not consider at the same time the individual variability of users and the characteristics of champions they select, therefore lacking contextual awareness. NICE stands out among these methods as a tool to not only predict outcomes but also to understand the underlying human behavior in an unsupervised manner.

\section{The League of Legends Dataset}
\subsection{Data Collection and Preprocessing} 
We use an LoL dataset collected from April 2014 to November 2016 \cite{sapienza2018individual}.  We preprocess the data to select only players who have a minimum of 15 matches within the time frame. First, this avoids statistical bias due to users who quit early on. Second, our method requires sufficient data points for each user so to obtain a meaningful profile of the user's playing habits. The final dataset we use consists of 18,713 players in 436,805 matches, totaling about 1 million unique player-match pairs.

\subsection{Feature Selection} 
The LoL dataset contains match metadata and individual player data. Match metadata includes the queue type, map ID, version number, season number as well as the date, time, duration, and the team-based outcome of the match (win or lose). There are also in-game performance statistics for each player, such as the number of \textit{kills}, \textit{deaths}, and \textit{assists}. The dataset also records each players' choice of the champion, the lane of the attack, and the role in the team, which we collectively deem as their individual contextual factors. Next, we briefly describe the meaning of the features we will be using. 

The queue type determines how the teams were formed. About half of the matches were solo-queue matches, meaning the teams comprised of solo players who are likely strangers with each other, as opposed queues where teams were partially or fully formed by the players themselves. About 60\% of the matches were draft queues, in which players can restrict the available champions the opposing team can pick for a more competitive game. Finally, almost all of the matches were ranked queues. The outcome of a ranked queue match goes towards the final ranking of a player at the end of a playing season, which is a yearlong process. In some game maps, players can play in 3v3 mode, but the 5v5 maps are more popular.

At the start of each match, players determine how they contribute to the overall layout of the team. This is a combination of choosing the lane on the map (e.g., top or bottom lane) and the role they play as part of the team (e.g., support or carry). Many of these choices conventionally depend on the champion they select: experienced players informally agree on the best lane or role combination that pairs with each champion, also known as the game's \textit{meta} or \textit{metagame} \cite{donaldson2017mechanics}. The 133 champions available during this time period are broadly categorized as one of \textit{Controllers} (17 champions), \textit{Fighters} (27), \textit{Mages} (23), \textit{Marksmen} (21), \textit{Slayers} (19), \textit{Tanks} (20) and \textit{Unique Playstyles} (6). Since the dataset offers only the champion name, we match champions with champion types according to the official list released in 2016.

Finally, the in-game performance measures include the number of \textit{kills}, \textit{deaths}, \textit{assists}, \textit{gold earned}, \textit{gold spent} and \textit{champion level} for each player in each match. Champion level is a score that a player accumulates throughout a match, with higher levels unlocking rewards and enhancing abilities. The champion level logged in the dataset is the final level attained at the end of the match. We additionally compute the \textit{kills-deaths-assist} (KDA) ratio, defined as $\textit{kills}+\textit{assists}/(\textit{deaths}+1)$, as an overall indicator of user performance. 

Like most MOBA games, LoL game developers closely monitor the game and regularly release patches about every two weeks. One of the main reasons for patches is to alter the dynamics of the game to ensure fair play. No single gameplay strategy or champion choice should dominate. Patches can ``nerf'' certain champions by decreasing the power level of their abilities and skills, or ``buff'' them through enhancements. Another reason for patches is to shake up the game in preparation for new seasons.  LoL seasons commonly last from January to November of each year and are followed by a short pre-season period. 
Pre-seasons are typically when the largest overhauls to maps, champions, and other game dynamics in the game are introduced. To incorporate game changes as features, we include the patch number as global contextual change points. There were 62 patches within the time frame of our dataset over from season 2014 to season 2016.

\subsection{Contextual Impact in LoL} \label{sec:context_impact}
Many in-game performance statistics are strongly related to the types of champions selected by a player in a match. Fig. \ref{fig:champ_plots} illustrates the performance measure trends for the seven types of champions. The differences in trends among champion types align with our intuition based on the characteristics by design: some champions are designed to be equipped with strong attacks, whereas others are mostly set up for positions of assist. For example, \textit{Slayer} champions, consistently ranked the highest in the number of kills, are known to have targeted, high bursts of damage. In contrast, \textit{Controller} champions output the least damage in terms of kills but outputs the highest level of assists, which allows them to assist teammates weakening their enemies and strengthening their teammates. Champion-based contextual information does not, however, directly impact winning outcomes. The concept of winning or losing is applied to the whole team, and teams customarily consist of a range of different champions and champion types.

Players tend to specialize in a few champions, and each champion they specialize in is usually only played in a specific role. While versatility in game-play can be a desirable attribute, many players choose instead to focus on practicing and polishing their skills on select champions. To confirm this, we measure the entropy of champion type distribution for each user. Suppose $P_i$ is the proportion of matches  user $x$ played as champion type $i$, then the champion type distribution entropy $H(x)$ of user $x$ is:
\begin{equation}
     H(x)=-\sum_{i}P_i\log P_i.
\end{equation}

\begin{figure}[t]
    \centering
    \includegraphics[width=\linewidth]{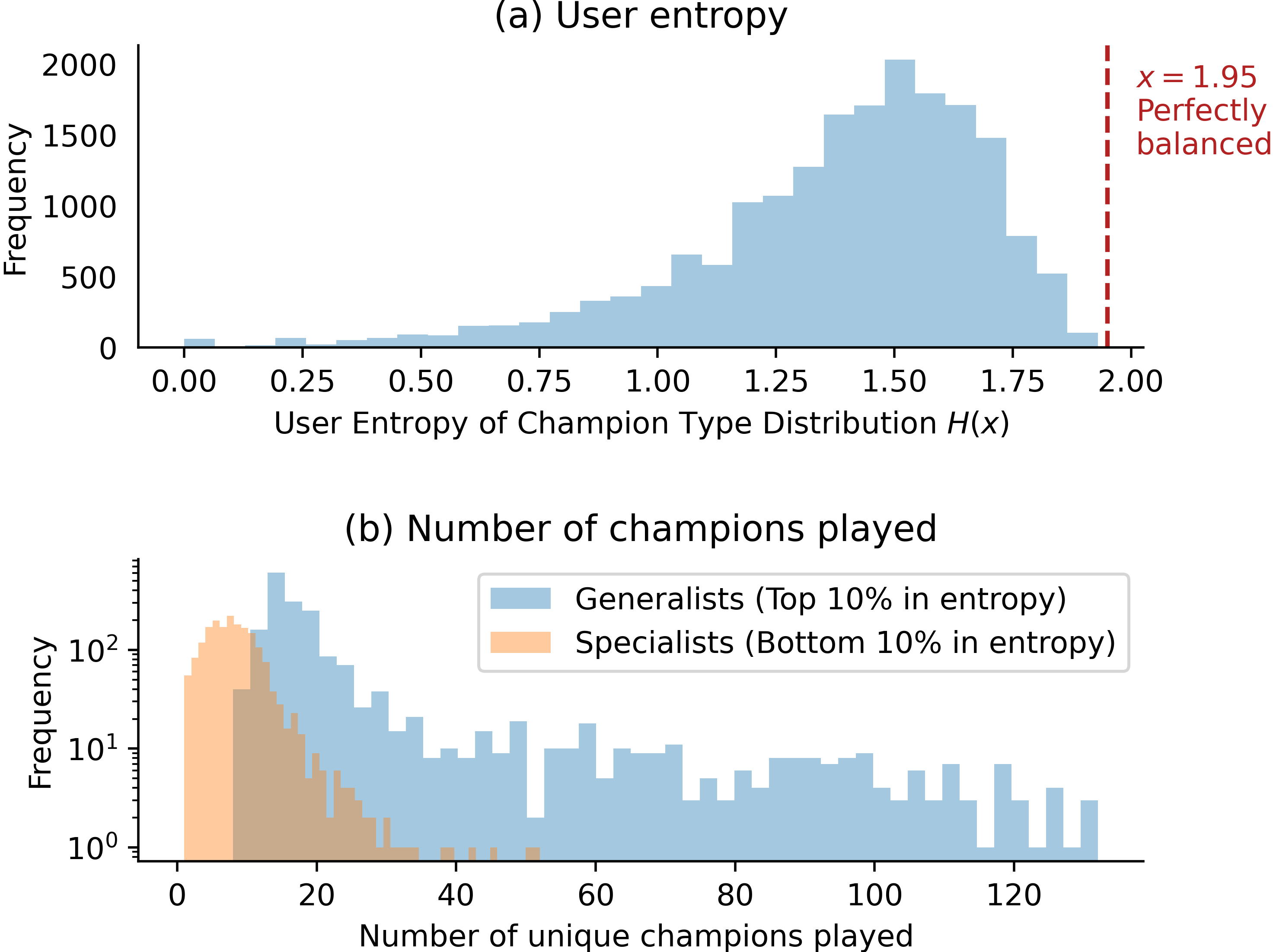}
    \caption{(a) Histogram of the user entropy of champion type distributions. The Perfectly Balanced bar indicates the entropy if a user played all seven types of champions equally.(b)  Histogram of the number of champions played per user for generalists and specialists. }
    \label{fig:entropy_dist}
\end{figure}

Fig. \ref{fig:entropy_dist}(a) displays a left-skewed distribution of entropy values. In particular, no user plays perfectly equally among all seven champion types, in which case the entropy would be maximal, i.e., equal to 1.95. We distinguish between \textit{generalists} -- who spread their usage more equally among champion types -- and \textit{specialists} -- who concentrate on fewer champion types -- at the top and bottom decile marks of entropy values, respectively. Most generalists spend more than 60\% of the time spread over 3 or more champion types, while specialists focus almost exclusively (over 50\% of the time) on one type of champion. In terms of unique champions, Fig. \ref{fig:entropy_dist}(b) shows that generalists are observed to have at least played 8 different champions at least once, with many exploring almost all 133 available champions. In contrast, most specialists play between 1 to 20 different champions. This reaffirms our belief that generalists and specialists exhibit fundamental differences in their gameplay preferences, supporting our hypothesis that a model accounting for the differences in user and distinctions among champions can lead to more accurate predictions.

\subsection{Prediction Tasks} Using the LoL dataset, we perform a range of prediction tasks including predicting the winning outcome, \textit{kills}, \textit{deaths}, \textit{assists}, KDA and user engagement. User engagement is measured by whether a match was the last match a player played in a continuous gaming session (i.e., an end-of-session match), where a session is defined as a series of consecutive matches divided by break intervals shorter than 15 minutes. Prior studies show that players' performance deteriorates throughout a MOBA game session \cite{sapienza2018individual} due to cognitive depletion. Thus, predicting the end of a session is helpful for understanding short term and long term engagement of the game. 

Depending on the prediction task, we exclude certain features from the training data due to dependency issues. For the target variables \textit{kills},  \textit{deaths}, and \textit{assists}, KDA is excluded. For the target variable KDA , \textit{kills}, \textit{deaths}, and \textit{assists} are excluded. 

\section{Methods}
\subsection{User-Context Tensor} 
We represent the data as a user-context tensor, which would be used in tensor factorization to mine context-dependent behavior patterns. Specifically, we consider two types of contexts:
\begin{enumerate}
    \item Global contexts: contexts that are applied to everyone;
    \item Individual contexts: contexts that are specific to each individual, either voluntarily or not.
\end{enumerate}
The resulting tensor is $\mathcal X$ is constructed as the $\textit{user}\times \textit{global  contexts}\times \textit{individual contexts}$ tensor, where \textit{users}, \textit{global contexts}, and \textit{individual contexts}  are each represented by unique identifiers. Each element $x_{ijk}\in \bm{\mathcal X}$ is the mean number of times user user $i$ experienced global context $j$ and individual context $k$. In our case, the global context is naturally the game version organized chronologically, which applies to every user simultaneously. Individual contexts are in-game context settings specific to each player, which we consider to be the champions. Thus, element $x_{ijk}\in \bm{\mathcal X}$ is the proportion of matches user $i$ played as champion $k$ in version $j$. Since feature dependencies should be avoided when using tensor factorization, we did not include context variables champion type, queue type, role, and lane in the tensor since they are strongly correlated with the choice of champions. Though our dataset contains only categorical context variables, our tensor construction method can be easily adapted to continuous context variables by taking the mean value instead of the normalized counts. Since not every user played the game during the time span of every version, the tensor is very sparse. Depending on the test ratio, only 20-30\% of the tensor is populated.
\begin{figure}[t]
    \includegraphics[width=\linewidth]{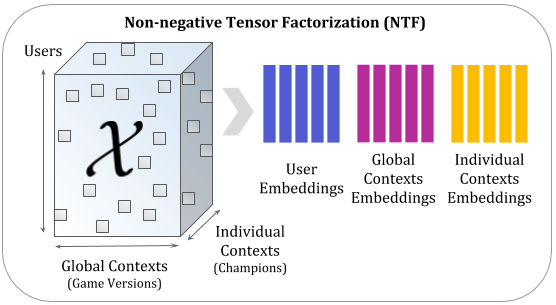}
    \caption{ Illustration of the user-context tensor of the LoL dataset and its factorized embeddings.}
    \label{fig:tensor}
\end{figure}
\subsection{Tensor Factorization} 
We decompose our data represented in tensor form using Non-Negative Tensor Factorization (NTF), a scalable algorithm based on PARAFAC/CANDECOMP, to obtain latent factors of $\bm{\mathcal X}$ \cite{kolda2009tensor}. These latent factors, or embeddings, are vector representations of the data in reduced dimensionality. By gleaning information from all users in various contextual settings, we can learn how different context settings impact each one individually. Fig. \ref{fig:tensor} shows an illustration of the user-context tensor when applied to the LoL dataset.

NTF works by approximating the tensor with a sum of rank-one tensors, subject to non-negativity constraints: 
\begin{equation}
    \bm{\mathcal X} \approx \sum_{r=1}^R \textbf u_r\circ \textbf t_r\circ\textbf f_r
\end{equation}
where $R$ is the chosen number of components and $\circ$ denotes outer product. This means that an element $x_{ijk}$ can be expressed as
\begin{equation}
    x_{ijk} \approx \sum_{r=1}^R u_{ir}t_{jr}f_{kr}
\end{equation}
for all $i\in1, ..., I$, $j\in 1, .., J$, and $k\in 1, ..., K$. We alternatively denote the latent factors as matrices of users $\textbf  U\in\mathbb R^{I\times R}$, versions $ \textbf T\in\mathbb R ^{J\times R}$ and champions $\textbf  F\in\mathbb R ^{K\times R}$, where $I$ is the number of users, $J$ is the number of versions, and $K$ is the number of champions. The tensor can be then concisely written as $ \bm{\mathcal X}  = [\![\textbf{ U}, \textbf{ T}, \textbf{ F}]\!]$ using the Kruskal operator \cite{kruskal1977three}. The $r$-th column of the matrices $\textbf U, \textbf T$ and $\textbf F$ corresponds to the column vectors $\textbf u_r, \textbf t_r$ and $\textbf f_r$, which we also call the \textit{component vectors}. The rows of each matrix correspond to a single user, version or champion depending on the factor matrix, which we also call the \textit{embedding} of a user, version or champion. The minimization problem in a typical CP decomposition problem is given by
\begin{equation}
    \begin{gathered}
    \min \bigg|\bigg|x_{ijk} - \sum_{m=1}^R\sum_{n=1}^R\sum_{l=1}^Ru_{im}t_{jn}f_{kl}\bigg|\bigg|^2_F\\
    \text{s.t.}\; u_{im}, t_{jn}, f_{kl}\geq 0 
    \end{gathered}
\end{equation}
where $||\cdot ||_F$ is the Frobenius norm. Due to the considerable sparseness in our tensor, performance will sharply degrade using standard practices for sparse data such as imputation. Instead, we use a masked version of the minimization problem:
\begin{equation}
    \min \bigg|\bigg|w_{ijk}\bigg(x_{ijk} - \sum_{m=1}^R\sum_{n=1}^R\sum_{l=1}^Ru_{im}t_{jn}f_{kl}\bigg)\bigg|\bigg|^2_F
\end{equation}
where $w_{ijk}$ is 0 if $x_{ijk}$ is missing and otherwise 1.
for all $i\in1, ..., I$, $j\in 1, .., J$, and $k\in 1, ..., K$. The problem is optimized using a nonlinear bound contrained optimization algorithm \cite{byrd1995limited}. For our work, we use the implementation of weighted CP decomposition for incomplete data publicly available in the Tensor Toolbox for MATLAB \cite{acar2011scalable}.

NTF depends heavily on the choice of the number of components. Too many components can lead to overfitting yet too few can lead to underfitting of the data
In many tensor factorization problems, tools such as the Core Consistency diagnostic test (CORCONDIA) \cite{bro2003new} or the ADD-ONE-UP method \cite{chen2001efficient} can be used to estimate an appropriate number of components. However, such methods will not work well in our case due to the large amount of missing data. Instead, we employ a cross-validation approach to select the most optimal number of components from the predictive results. We detail our experimental setup in Section \ref{sec:results1}.

\subsection{Deep Neural Network Decoder}
With the latent representation of users, global contexts, and individual features, we make predictions using a deep neural network decoder. At this stage, a training instance is a unique user-match pair. 

\begin{table*}[]
    \centering
    \caption{Prediction Results}
    \begin{threeparttable}
    \begin{tabular}{rcccccccccc}
        \toprule 
         &Winning Outcome&End of Session & \multicolumn{2}{c}{KDA} &\multicolumn{2}{c}{Kills} & \multicolumn{2}{c}{Deaths} & \multicolumn{2}{c}{Assists}  \\
         &AUC & AUC & RMSE & NRMSE & RMSE & NRMSE & RMSE & NRMSE & RMSE & NRMSE \\
         \midrule
         Random Forest & 0.920 & 0.532 & 2.161& 0.048 &  2.401& 0.057 &2.534 & 0.050 & 3.844& 0.078\\
         XGBoost & 0.948 & 0.579 & 1.982 & 0.044 & 2.216& 0.052 & 2.361 & 0.046 & 3.563& 0.073 \\
         DNN &  0.945 & 0.576 &  1.978   & 0.044 & 2.215  & 0.053 & 2.366 & 0.046 & 3.563 & 0.073 \\
         NICE & \textbf{0.953} & \textbf{0.680} &\textbf{1.934} & \textbf{0.043} &\textbf{2.008} & \textbf{0.048} & \textbf{2.299} & \textbf{0.045} & \textbf{3.367}& \textbf{0.069} \\ 
         \bottomrule
    \end{tabular}
    \begin{tablenotes}
    \item 
    We evaluate classification problems with AUC (the higher the better) and regression problems with RMSE and NRMSE (the lower the better). The best results are indicated in bold.
    \end{tablenotes}
    \end{threeparttable}
    \label{tab:results}
\end{table*}

Let $\textbf x_i^f$ denote the vector of individual context (champion types) $i$.  For simplicity, we will drop the subscript $i$ whenever it is otherwise clear. We integrate $\textbf x^f$ with the learned individual context embeddings $\textbf F$ as follows:
\begin{align}
    \textbf F_i'= \textbf x^f \odot \textbf f\; \text{with } F'_{jk} = x^f_jF_{jk},
\end{align}
where $\odot$ is the Hadamard, or element-wise, product. We can understand $\textbf F'$ to be the individual context matrix for a user-match instance. $\textbf F'$ is reduced to a 1-dimensional vector $\textbf f\in\mathbb R^R$ by summing up the values at every component $r$. Since each user-match instance only has one champion type, $\textbf x_i^f$  is a binary vector with a 1 for the selected champion and 0 anywhere else. The resulting individual context matrix $\textbf F'$ has all but one row of zeros for our specific problem. However, this framework works for any format of individual contexts.

To incorporate $\textbf f_i$ with other features and embeddings, we create a concatenated vector $\textbf h_i=[\textbf u_i,\textbf f_i,\textbf t_i,\textbf x_i^m]$, where $\textbf u_i$ is the user embedding, $\textbf t_i$ is version embedding, and $\textbf x_i^m$ is the set of all other features in training instance $i$. Other features refer to the season, queue type, champion type, role, lane, map ID, match duration, and the timestamp of the match. Also potentially included in $\textbf x_i^m$  are \textit{kills}, \textit{deaths}, \textit{assists}, and KDA, provided they are not the target variable or not affiliated with the target variable. All continuous features are normalized and all categorical features are one-hot encoded. $\textbf h_i$ is then sent to 8 fully connected layers with hidden sizes ranging from 258 to 2 in decreasing powers of 2. Every layer is followed by a Leaky ReLU activation function and optionally a Dropout \cite{srivastava2014dropout} layer.

The output layer of the neural network produces a singleton, applied with either a ReLU (for regression) or sigmoid (for binary classification) activation function. 
Depending on the task, we minimize either the mean squared error (regression) or the binary cross-entropy (binary classification) with the Adam optimizer \cite{kingma2014adam}. 
All weights are regularized with a beta of $1\times 10^{-7}$. We train the data on a batch size of 2048 until convergence using a learning rate of $1\times 10^{-3}$.

\section{Results}

\subsection{Evaluation}\label{sec:results1}

A random sample of 20\% of the data is held-out as the test set. This split is stratified by user ID, as NICE must learn user embeddings for every user. The remainder of the 80\% of the data is used for training and hyperparameter tuning. 

Baseline comparisons selected are Random Forest \cite{liaw2002classification} and XGBoost  \cite{chen2016xgboost} for their adaptability to both regression and classification tasks and also for their ease of scaling up to large input data. Both baseline models are grid searched for the most optimal hyperparameter using 5-fold cross-validation independently for each target variable. Additionally, to show the effectiveness of using embeddings as latent features, we use the same DNN decoder built for NICE as another baseline model by not providing the DNN the latent embeddings as feature inputs. Instead of latent embeddings, all baseline models receive user IDs and champion IDs as one-hot encoded vectors and the version number as a discrete numeric feature. The rest of the feature inputs remain the same. 

Binary classification problems are evaluated using AUC and the regression problems are evaluated using RMSE. To facilitate the comparison of results for regression target variables that span different ranges, we also compute the Normalized NRMSE:
\begin{equation}
    \textit{NRMSE} = \frac{\textit{RMSE}}{Y_{\max} -
    Y_{\min}} = \frac{\sqrt{\frac{1}{n}\sum_{i=1}^n(\hat y_i - \hat y)}}{Y_{\max} - Y_{\min}},
\end{equation}
NRMSE can be used to compare the results of various regression tasks on the same scale. 

The optimal number of components used in the tensor factorization step is the most crucial hyperparameter for the NICE models. We explore the predictive performance of various component numbers from 1 to 10. We determined the most optimal number to be 6, the smallest number that produces the best result for all target variables without the risk of overfitting. 

Table \ref{tab:results} reports our experimental results for each target variable. Examining the results, we observe that almost all baselines perform similarly, with XGBoost and DNN being the best performing baseline models. Importantly, NICE delivers improvement across all prediction tasks. The prediction of winning outcome appears to be an easy task, with even baseline models achieving AUCs as high as .948. For this task, NICE outperforms the baselines with an improvement of .953 AUC. For the problem of predicting the end of a session, NICE significantly increases the AUC by 17\%, which may be because some players are more likely to play longer sessions in one sitting than others, a behavioral pattern that is picked up by NICE. As for the regression tasks of predicting \textit{kills}, \textit{deaths}, \textit{assists}, and \textit{KDA}, NICE produces better results than baselines for all target variables. Of the four, \textit{kills} and \textit{assists} are more challenging to predict due to their smaller baseline NRMSEs. It is also for these two target variables we see the biggest improvement using NICE, with a 9\% gain in NRMSE from the next best model for \textit{kills} and a 6\% for \textit{assists}. For target variables \textit{KDA} and \textit{Deaths}, NICE raises the baseline NRMSE by .001 ($\approx$ 2\%). Revisiting Fig. \ref{fig:champ_plots}, we see that the trends in \textit{kills} and \textit{assists} are more clearly segmented by champion types than \textit{deaths} and \textit{KDA}, explaining why NICE is especially good at predict \textit{kills} and \textit{deaths} in the context of champion selection. These results support as evidence that the contextually-aware model NICE leads to more informed predictions.

\begin{table}[]
    \begin{threeparttable}
    \caption{Number of users and champions with component labels}
    \setlength{\tabcolsep}{4pt}
    \begin{tabular}{r|ccccccc}
    \toprule
    &\multicolumn{6}{c}{Component Labels}& \multirow{2}{*}{Unlabeled}\\
    & 0 & 1 & 2 &3 &4&5&\\
    \midrule 
     Users & 176 &1148 & 2659 & 2495 &2386 & 2476 & 7373\\
     Champions  & 15 & 16 & 18 & 31 & 20 & 38 & 4 \\
     \bottomrule
    \end{tabular}
    \begin{tablenotes}
    \item user or champion is assigned component label $i$ if the $i$th component of its normalized embedding is the maximally activated component and at least 0.4.
    \end{tablenotes}
    \end{threeparttable}
    \label{tab:user_champion_labels}
\end{table}

\begin{figure*}
    \centering
    \includegraphics[width=0.8\linewidth]{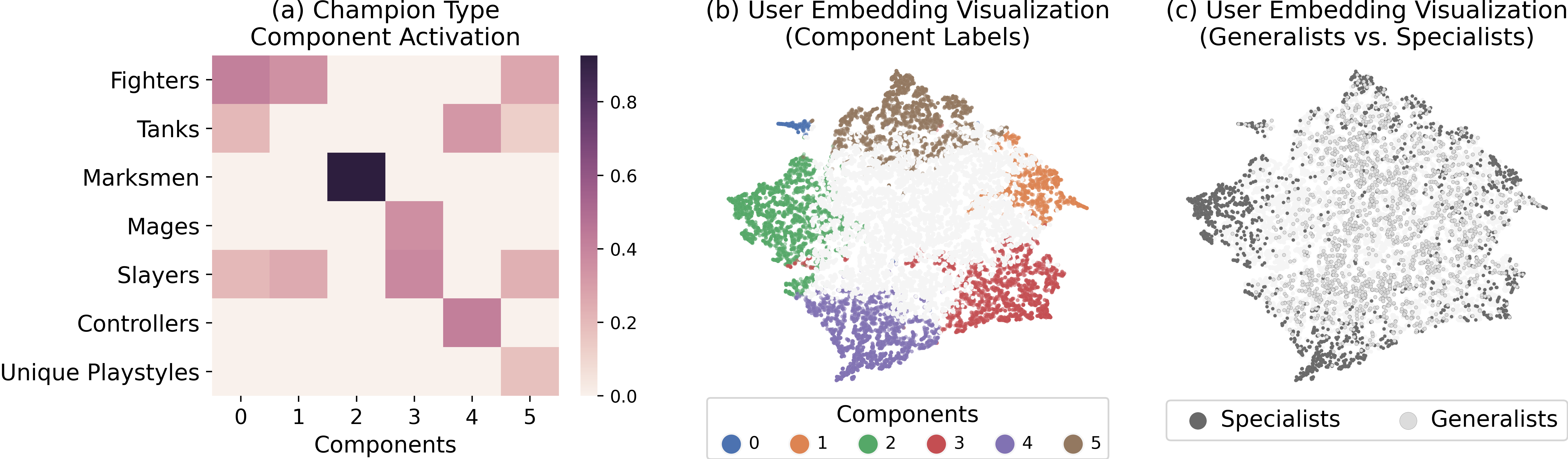}
    \caption{(a) Component activation of champion embeddings aggregated by their champion types. For each row of the seven champion types, we sort entries in descending order and then mask out any entries that are greater than 0.95 of the cumulative sum. (b) TSNE visualization of the user embeddings, where users assigned component labels are color-coded accordingly, and users without component labels are shown in white. (c) TSNE visualization of specialists and generalists users.}
    \label{fig:champion_tsne}
\end{figure*}
\subsection{Interpretation of Embeddings}
A unique advantage of using tensor factorization is its interpretability. For this section, we apply NTF  ($R=6$) on the entire dataset to visually analyze the meaning of the factorized matrices. 

We first identify the component membership level of champions types. This is done by first averaging champion embeddings to produce champion type embeddings. All entries are then normalized by component so that each component sums up to 1. We then rank the square of the entries in decreasing order and zero out entries that are beyond 95\% of the norm of the component. From \ref{fig:champion_tsne}(a) we see that, on average, each component identifies with 2 to 3 champion types, with \textit{Marksmen} champions being strongly identified by a single component.
\begin{figure*}
    \centering
    \includegraphics[width=\linewidth]{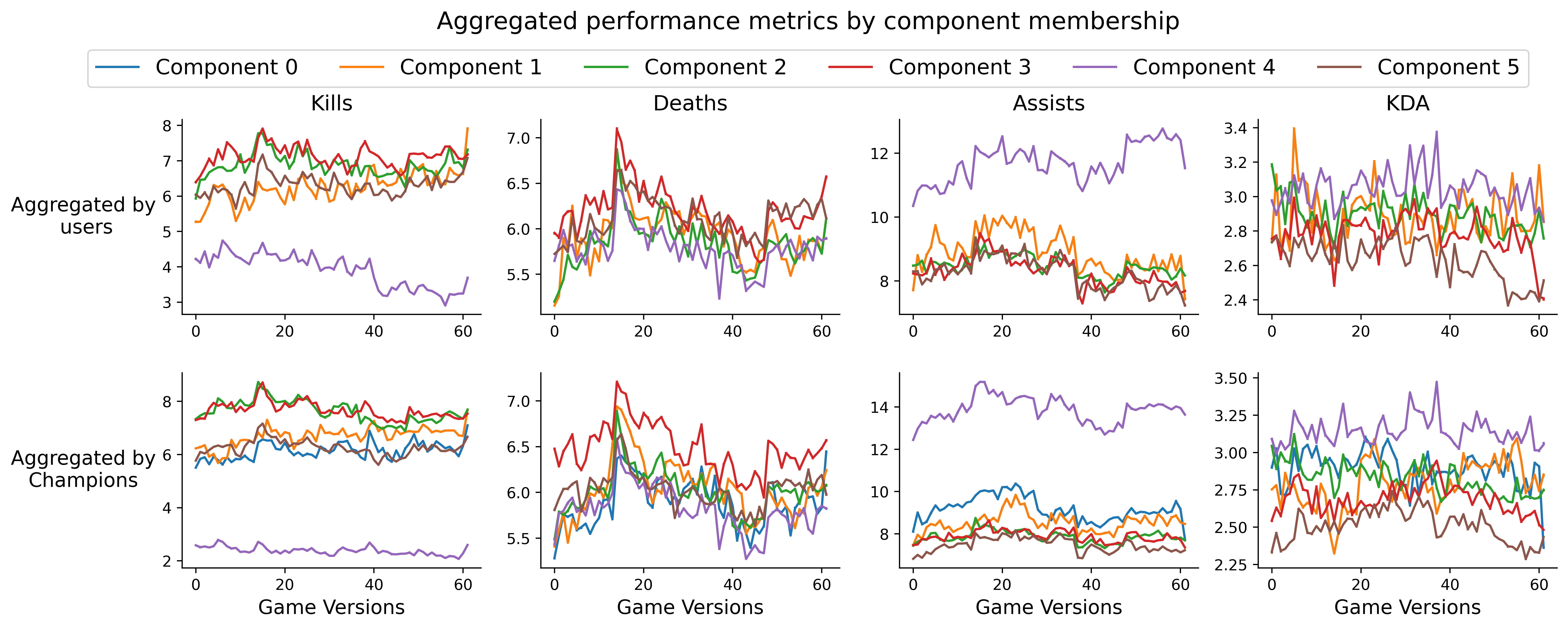}
    \caption{Time series of in-game performance measures for data points for which the users (top row) or champions (bottom row) are assigned a maximally activated component label. Component 0 is omitted in the user label aggregation due to the small number of users labeled as component 0, leading to high fluctuations in performance trends that distract visibility. }
    \label{fig:user_champion_perf}
\end{figure*}

To enhance the interpretability of the user factorization matrix $\bm{ U}$, we assign the component that each user most prominently identifies with as their label. We only assign labels to those whose maximally activated component contributes at least 40\% to the total activation of the user's embedding. The number of users labeled as each component is listed in Table \ref{tab:user_champion_labels}. Component label 0 contains the fewest number of people, having less than 200, while all other component labels contain at least 1000 users. Less than half of the users have relatively evenly activated embeddings and therefore was not assigned a component label. The labeled user embeddings are visualized in Fig. \ref{fig:champion_tsne}(b) using TSNE \cite{maaten2008visualizing}. Users sharing component labels occupy one cluster, indicating the embeddings of users who share the same maximally activated component are also overall similar to each other. In Fig. \ref{fig:champion_tsne}(c), we plot only the generalists and the specialists that determined from Section \ref{sec:context_impact} based on their champion type distributions. Specialists are concentrated in the various corners of the plot, whereas generalists, having more balanced activation, are scattered in the middle. This shows that the user embeddings indeed capture characteristics of specialization, or lack thereof, of some users.

We assess how components are related to in-game performance measures. This can be done by tracking users identified most prominently by each component label or by tracking champions using the same method (Table \ref{tab:user_champion_labels}). We show the time series of performance measures across game versions aggregated by user labels or champion labels (Fig. \ref{fig:user_champion_perf}). The layered pattern across all performance measures according to component labels is visually similar to the performance trends by champion types in Fig. \ref{fig:champ_plots}: each component can be ranked according to their average performance measure, and this ranking is stable over time. Moreover, the time-series of performance measures partitioned by users and by champions exhibit parallel trajectories, a phenomenon that may be partly explained by the fact that users with particular component labels are frequent players of the champions of the same labels. Some components immediately stand out as having distinct characteristics in performances. For example, component 4 (purple) is characterized by low \textit{kills}, and high \textit{assists}. From Fig. \ref{fig:champion_tsne}(a) we know that component 4 is primarily associated with the champion class \textit{Controllers} and these characteristics are indeed consistent with \textit{Controller} performance trends (Fig. \ref{fig:champ_plots}). Similarly, component 3 (red), associated with Mages and Slayers (Fig. \ref{fig:champion_tsne}(a)), is characterized by high \textit{kills}, high \textit{deaths} and low \textit{assists}, a pattern in line with champion types Mages and Slayers from Fig. \ref{fig:champ_plots}.

\begin{figure}[t]
    \centering
    \includegraphics[width=\linewidth]{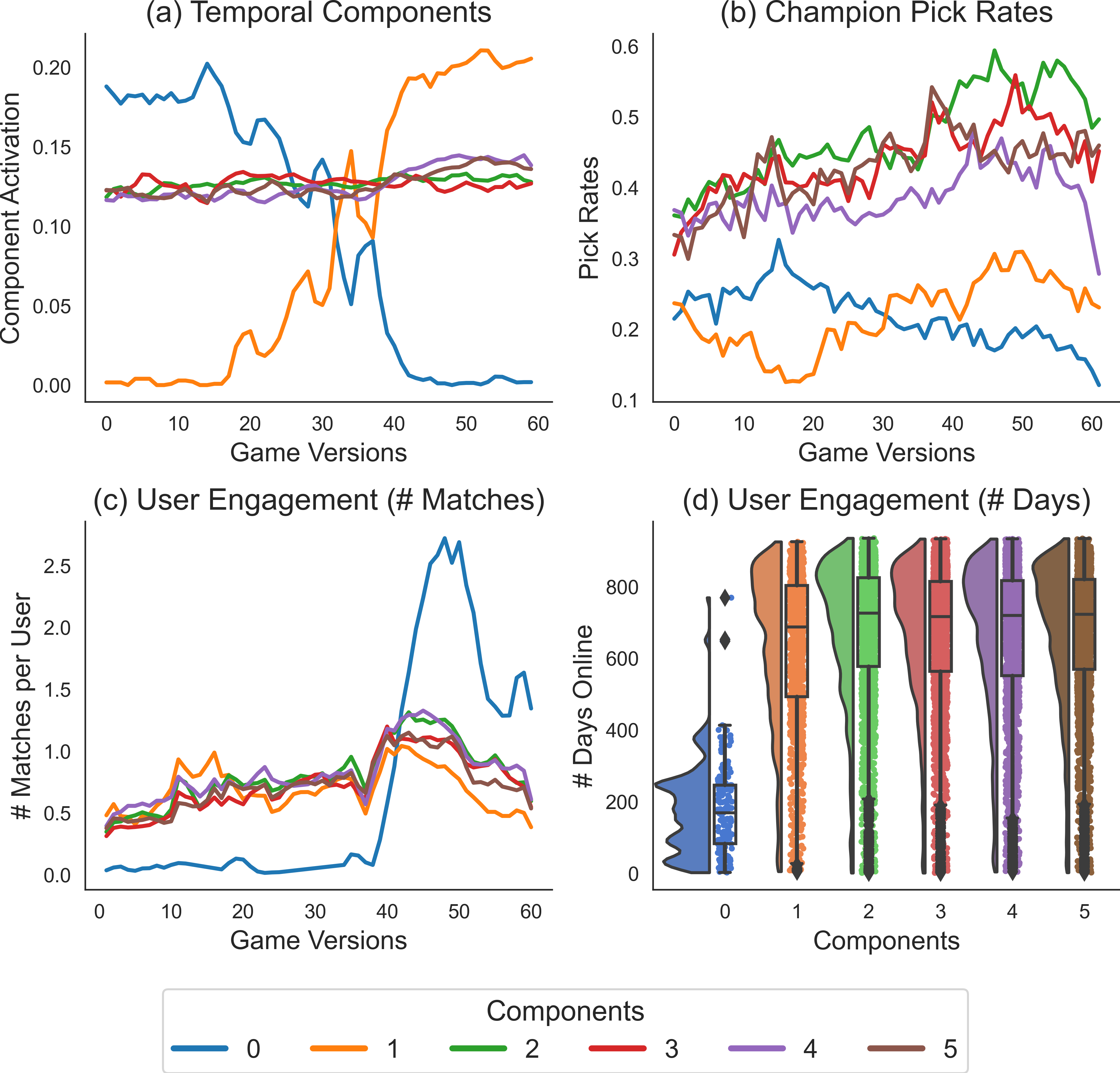}
    \caption{(a) 3-unit moving average of the temporal component activation. (b) Pick rates of champions under each component label. (c) User engagement measured in terms of the average number of matches per user in each game version for users under each component label. (d) A Raincloud plot \cite{allen2019raincloud} of user engagement measured in terms of the number of days users were online.}
    \label{fig:temporal}
\end{figure}

Finally, the interpretation of the temporal factor matrix (Fig. \ref{fig:temporal}(a)) offers a few additional insights particularly for component 0 and its relation to the year 2016. While the temporal components 2 through 5 remain fairly stable over time, components 0 and 1 demonstrate an alternating pattern intersecting at game versions indexed 36-40, which happens to be pre-season 2016. At this time, the champion pick rates of champions, which is the number of matches a champion was picked to play out of the number of total matches, exhibit similar patterns. Indeed, the pick rates and the temporal activation correlate with a Pearson's correlation coefficient of 0.88 and 0.83 for components 0 and 1, respectively, suggesting that component vectors 0 and 1 correspond to two distinct groups of champions. Component 0 tracks champions that were more popular before pre-season 2016 and component 1 tracks champions that only grew in popularity starting from pre-season 2016. 

Fig. \ref{fig:temporal}(c)-(d) also illustrates a relationship between component 0 and user engagement. Beginning from game version 40 (the start of season 2016), the engagement from this group of users increased drastically and is higher than the average engagement of other users. Fig \ref{fig:temporal}(d) depicts a Raincloud plot \cite{allen2019raincloud} of the number of days online recorded for each user group. In stark contrast to more than 600 average active days from users in other components, component 0 comprises a comparatively much more novice group of users, averaging less than 200 active days, suggesting that component 0 corresponds to a group of novice players dedicated to playing LoL in season 2016.

\section{Discussion}
Using tensor factorization methods on user-context tensor, our model NICE considerably improves prediction over baseline methods that are applicable and amenable to similar types of problems. The advantage of NICE is its personalized, context-aware deep learning approach. By drawing information from all users and all contexts, NICE learns from the dataset holistically, rather than locally at each data point. Further, while we applied NICE on an LoL dataset, we also describe how NICE can be flexibly applied to any human-centered large datasets in which there are contextual variables.  User-context can be formed in situations in which there are (i) global contexts that affect the population and (ii) individual contexts specific to each user. The same principals can be even more easily applied when there are only individual-specific contexts using matrix factorization. 

In addition to the contextually informed predictions, NICE is highly interpretable. The factor matrices induced by the NTF procedure opens the opportunity for many human behavior analyses, such as examining the effects of contextual changes or treatments on users. The underlying heterogeneity of users as observed by their chosen in-game contexts and their given global contexts is encapsulated in the latent embeddings. 

As for future work, our effort will be two-pronged: from a methodological standpoint, we will work to extend our methodology to even richer, multi-way data; from what concerns empirical applications, we will investigate the use of our framework to predict the performance or behavior of individuals in a variety of settings, from personalized health to social media. In all these domains, user-context data is abundant. A framework like ours can be beneficial for revealing hidden patterns in behavior, which can not only improve prediction but also help explain and interpret the data.

\section*{Acknowledgments} 
The authors are grateful to DARPA for support under AIE Opportunity No DARPA-PA-18-02-07. This project does not necessarily reflect the position or policy of the Government; no official endorsement should be inferred.



%

\bibliography{bib.bib}{}
\bibliographystyle{IEEEtran}

\end{document}